\documentclass[letterpaper]{article} 
\usepackage{aaai24}  
\usepackage{times}  
\usepackage{helvet}  
\usepackage{courier}  
\usepackage[hyphens]{url}  
\usepackage{graphicx} 
\usepackage{natbib}  
\usepackage{caption} 
\usepackage{graphicx}
\usepackage{textcomp}
\usepackage{subfigure}
\usepackage{url}
\usepackage{verbatim}
\usepackage{graphicx}
\usepackage{amssymb}
\usepackage{amsmath}
\usepackage{array,booktabs}
\usepackage{multirow}
\usepackage{bm}
\usepackage{bbding}
\usepackage{pifont}
\usepackage{makecell}
\usepackage{xcolor}
\usepackage{array}
\usepackage{tabularx}
\usepackage{newfloat}
\usepackage{listings}
\usepackage{algorithm}
\usepackage{algorithmic}

\usepackage{newfloat}
\usepackage{listings}
\DeclareCaptionStyle{ruled}{labelfont=normalfont,labelsep=colon,strut=off} 
\floatstyle{ruled}
\newfloat{listing}{tb}{lst}{}
\floatname{listing}{Listing}
\title{BAT: Behavior-Aware Human-Like Trajectory Prediction for Autonomous Driving}
\author {
    Haicheng Liao \textsuperscript{\rm 1}\thanks{Authors contributed equally; \dag Corresponding author.}, Zhenning Li\textsuperscript{\rm 1}$^{*\dag}$, Huanming Shen\textsuperscript{\rm 2}, Wenxuan Zeng\textsuperscript{\rm 3}, Dongping Liao\textsuperscript{\rm 1}, Guofa Li\textsuperscript{\rm 4}, Shengbo Eben Li\textsuperscript{\rm 5}, Chengzhong Xu\textsuperscript{\rm 1}$^{\dag}$
}
\affiliations{
    \textsuperscript{\rm 1} University of Macau\\
    \textsuperscript{\rm 2} University of Electronic Science and Technology of China\\
    \textsuperscript{\rm 3} Peking University\\
    \textsuperscript{\rm 4} Chongqing University\\
    \textsuperscript{\rm 5} Tsinghua University\\
    \{yc27979, zhenningli, yb97428, czxu\}@um.edu.mo, huanmingshen@outlook.com,\\ zwx.andy@stu.pku.edu.cn, hanshan198@gmail.com, lishbo@tsinghua.edu.cn
}
\usepackage{bibentry}

\begin{document}

\maketitle

\begin{abstract}
The ability to accurately predict the trajectory of surrounding vehicles is a critical hurdle to overcome on the journey to fully autonomous vehicles. To address this challenge, we pioneer a novel behavior-aware trajectory prediction model (BAT) that incorporates insights and findings from traffic psychology, human behavior, and decision-making. Our model consists of behavior-aware, interaction-aware, priority-aware, and position-aware modules that perceive and understand the underlying interactions and account for uncertainty and variability in prediction, enabling higher-level learning and flexibility without rigid categorization of driving behavior. Importantly, this approach eliminates the need for manual labeling in the training process and addresses the challenges of non-continuous behavior labeling and the selection of appropriate time windows. We evaluate BAT's performance across the Next Generation Simulation (NGSIM), Highway Drone (HighD), Roundabout Drone (RounD), and Macao Connected Autonomous Driving (MoCAD) datasets, showcasing its superiority over prevailing state-of-the-art (SOTA) benchmarks in terms of prediction accuracy and efficiency. Remarkably, even when trained on reduced portions of the training data (25\%), our model outperforms most of the baselines, demonstrating its robustness and efficiency in predicting vehicle trajectories, and the potential to reduce the amount of data required to train autonomous vehicles, especially in corner cases. In conclusion, the behavior-aware model represents a significant advancement in the development of autonomous vehicles capable of predicting trajectories with the same level of proficiency as human drivers. The project page is available at our Github\footnote{{https://github.com/Petrichor625/BATraj-Behavior-aware-Model}}. 
\end{abstract}

\section{Introduction}
\begin{figure}[t]
  \centering  \includegraphics[width=\linewidth]{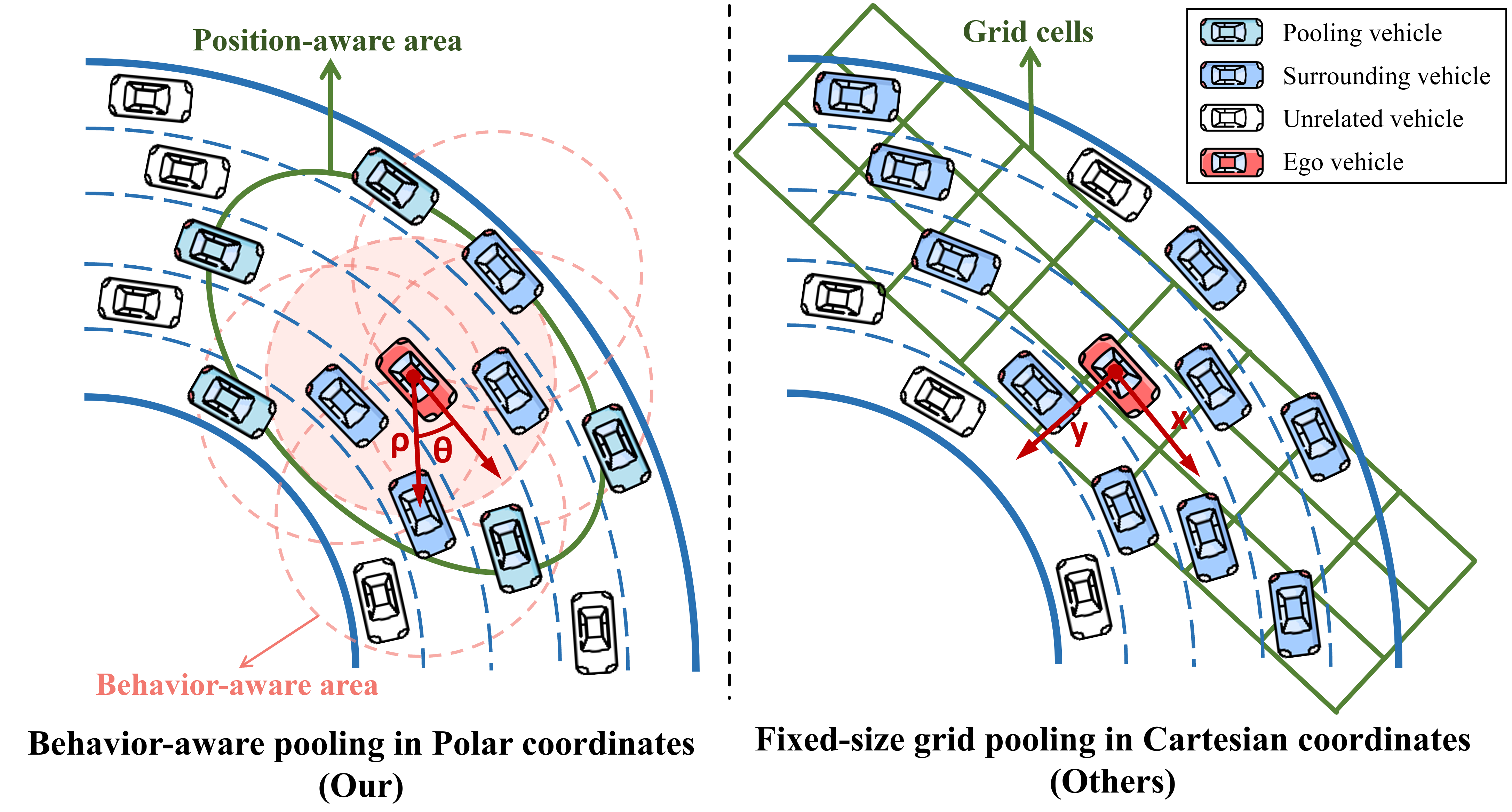} 
  \caption{An overview of our proposed behavior-aware pooling mechanism and the classical pooling mechanism. Left: Modeling the vehicle using polar coordinates. Right: Modeling the vehicle using fixed-size grids and representing the position in Cartesian coordinates.}
  \label{polar} 
\end{figure}
Recent advancements in autonomous driving (AD) have been remarkable. Nonetheless, as we move towards the commercialization of high-level AD technology, challenges abound. One of the most significant barriers is equipping autonomous vehicles (AVs) with the ability to anticipate the trajectory of nearby vehicles in intricate situations as skillfully as humans.

Driving, for humans, necessitates continuous monitoring of the current states of surrounding vehicles and forecasting their future states before actions like acceleration or overtaking. These states, predominantly determined by trajectories, form the bedrock of safe driving and collision prevention. This demands a keen assessment of the \textit{interaction} among vehicles and an unbiased grasp of their \textit{behavior}, in line with traffic regulations and accumulated driving experience \cite{muller2016social}.

In our quest to enhance the trajectory prediction capabilities of AVs, mimicking human-like comprehension and response to surrounding scenarios might be a breakthrough. As highlighted in \citep{schwarting2019social,wang2022social}, accounting for the behaviors of other drivers in the decision-making processes of AVs can potentially result in enhanced driving performance. With this understanding, we advocate that a deeper dive into driver behavior can significantly uplift trajectory prediction for AVs.

Previous investigations have posited that there exists a certain relationship between different drivers' behaviors and their driving performance \citep{toledo2008vehicle,chandra2020cmetric,xie2020motion}. When confronted with the prospect of another vehicle attempting to overtake, aggressive drivers may accelerate to impede the overtaking vehicle, while cautious drivers may reduce their speed slightly to facilitate safe passing.  In addition, driver behavior on the road tends to exhibit a degree of predictability, persistence, and consistency \citep{hang2022human,schwarting2019social}. For example, an individual who has recently violated the speed limit is likely to continue driving at high speeds as long as circumstances allow, while cautious drivers maintain their conservative driving strategy. These stability and repetition characteristics make it possible to predict and anticipate the behavior of other drivers. 

In addition, humans naturally perceive their surroundings in relative terms, especially when it involves spatial understanding. This intrinsic way of processing spatial data based on relative positioning and orientation often does not align with the fixed Cartesian coordinates commonly used in many predictive models. However, polar coordinates, which detail a point's position based on its distance from a reference and the angle from a reference direction, echo this human-centric perception. When driving, humans think in terms like "slightly ahead and to the right" rather than specific Cartesian coordinates. Adopting this perspective, our pioneering pooling mechanism, as illustrated in Fig.1, captures vehicle positions using polar coordinates, offering a more intuitive representation especially pertinent for trajectory prediction in AVs. 

Despite extensive research in AD trajectory prediction, significant gaps remain. To bridge these, we've combined insights from human behavior and decision-making to design an innovative behavior-aware trajectory prediction model. In summary, our work's principal contributions are:
\begin{itemize}
    \item  We present a novel dynamic geometric graph approach that eliminates the need for manual labeling during training. This method addresses the challenges of labeling non-continuous behaviors and selecting appropriate time windows, while effectively capturing continuous driving behavior. Inspired by traffic psychology, decision theory, and driving dynamics, our model incorporates centrality metrics and behavior-aware criteria to provide enhanced flexibility and accuracy in representing driving behavior. To the best of our knowledge, this is the first attempt to incorporate \textbf{continuous representation of behavioral knowledge} in trajectory prediction for AVs.
    \item We propose a novel pooling mechanism, aligned with human observational instincts, that extracts vehicle positions in \textbf{polar coordinates}. It simplifies the representation of direction and distance in Cartesian coordinates, accounts for road curvature, and allows modeling in complex scenarios such as roundabouts and intersections.
    \item We introduce a new Macao Connected Autonomous Driving (MoCAD) dataset, sourced from a L5 autonomous bus with over 300 hours across campus and busy urban routes. Characterized by its unique \textbf{right-hand-drive system}, MoCAD, set to be publicly available, is pivotal for research in right-hand-drive dynamics and enhancing trajectory prediction models.
    \item Our model significantly outperforms the SOTA baseline models when tested on the NGSIM, HighD, RounD, and MoCAD datasets, respectively. Remarkably, it maintains impressive performance even when trained on only 25.0\% of the dataset, demonstrating exceptional robustness and adaptability in various traffic scenarios, including \textbf{highways}, \textbf{roundabouts}, and \textbf{busy urban locales}. 
\end{itemize}

\section{Related Work}\label{Related work}
A plethora of research has been conducted in the realm of trajectory prediction, with a diverse array of approaches being proposed. These approaches can be broadly classified into three categories: physics-based, statistics-based, and deep learning-based approaches. 

\textbf{Physics-based Approaches.}
These approaches are primarily divided into kinetic and kinematic models \cite{lin2000vehicle}. They use principles from physics and mechanics, taking into account the current state of the vehicle, such as speed and steering angle, to make predictions \cite{batz2009recognition,wong2022view}. Despite their interpretability and computational efficiency, these methods often exhibit lower prediction accuracy compared to SOTA techniques \cite{huang2022survey}.

\textbf{Statistics-based Approaches.}
In contrast, statistical-based approaches, both parametric and non-parametric, describe predicted trajectories using predefined maneuver distributions, such as Gaussian processes, hidden Markov models, dynamic Bayesian networks, and support vector machines \cite{wang2021decision,li2020pedestrian,xie2017vehicle,li2023mitigating}. These methods tend to offer more refined and sophisticated model structures, resulting in better prediction performance than physics-based approaches. Their experiments on real-world data showed significant improvements over baselines.

\textbf{Deep Learning-based Approaches.} 
The surge in popularity of deep learning has led to extensive research in trajectory prediction for AVs. 
Recurrent Neural Networks (RNNs), Convolutional Neural Networks (CNNs), and Transformers \cite{vaswani2017attention} are among the most widely used approaches, each offering unique modeling considerations and focuses \cite{ye2021tpcn,liang2020learning}. RNNs, such as Long Short-Term Memory (LSTM), are often used to process time-series trajectory data, while CNNs excel at extracting spatial features from inputs such as bird's-eye or raster images. Some researchers combine RNNs and CNNs to integrate both temporal and spatial features into their models \cite{liao2023gpt,9812060,bhattacharyya2023ssl, zhang2022explainable}.  Transformers, with their renowned success in many domains, have also demonstrated superior performance in trajectory prediction \cite{li2022lane,zeng2023mpcvit,li2023context}. Compared to physics-based and statistics-based methods, these data-driven approaches have generally demonstrated superior prediction performance, especially for tasks requiring long-term predictions (beyond 3 seconds).

\section{Problem Formulation}\label{Problem Formulation}
The trajectory prediction task can be formulated as follows. At each time $t$, we predict the multimodal trajectories of the ego vehicle, based on historical observations of both the ego vehicle and its surrounding vehicles (\textit{agents}). Given the inputs of historical observations $\bm{X}$, the model aims to predict a multi-modal distribution over future trajectories of the ego vehicle $P(\bm{Y}|\bm{X})$. 

\subsection{Inputs and Outputs}
The inputs $\bm{X}$ to our model are the historical trajectories over a fixed time horizon $t_{h}$ of both the ego vehicle (subscript $0$) and all the surrounding vehicles (subscripts $1$ to $n$):
\begin{equation}\label{eq.1}
   \bm{X}_{i}^{t-t_{h}:t} = \left\{p_{i}^{t-t_{h}:t} \right\}, \forall i\in[0,n]
\end{equation}
where $ p_{0:n}^{t-t_{h}:t}$ denotes the 2D position coordinates.

The output of the model is a probability distribution over the future trajectory of the ego vehicle during the prediction horizon $t_{f}$ :
\begin{equation}\label{eq.4}
    \bm{Y} = \bm{Y}_{0}^{t+1:t+t_{f}}= {\{y_{0}^{t+1},y_{0}^{t+2},\ldots,y_{0}^{t+t_{f}-1},y_{0}^{t+t_{f}}\}}
\end{equation}

As aforementioned, we define the motion of the vehicles in Polar coordinates (shown in Fig.\ref{polar}) rather than the Cartesian coordinates. In the Polar coordinate, we assume the origin $ {O}$ of the stationary frame of reference is fixed at the center of the ego vehicle at time $t$. Our inputs and outputs can be further written as (take the ego vehicle as an example, for convenience, assume input at instant $t$ and output at instant $t+1$): 
\begin{equation}
    x_{0}^{t}= \{\rho_{0}^{t},\theta_{0}^{t}\}
\end{equation}
and
\begin{equation}
    y_{0}^{t+1}= \{\rho_{0}^{t+1},\theta_{0}^{t+1}\}
\end{equation}
where $\rho$ and $\theta$ are the distance and angle of the vehicle. 

The transformation relationship between Cartesian and Polar coordinate systems is illustrated below. Given a vehicle’s position history in lateral coordinate $x_i^{t_{k}}$ and longitudinal coordinate $y_i^{t_{k}}$ at time $t_{k}$, the distance $\rho_i^{t_{k}}$ and vehicle orientation $\theta_i^{t_{k}}$ for Polar representation can be computed as the following formula:
\begin{equation}\label{eq.-1}
    \left\{\begin{array}{l}
    \rho_i^{t_{k}}=\sqrt{\left(x_{i}^{t_{k}}-x_{0}^{t}\right)^2+\left(y_{i}^{t_{k}}-y_{0}^{t}\right)^2} \\
    \theta_{i}^{t_{k}}=\arctan \left(\frac{y_{i}^{t_{k}}-y_{0}^{t}}{x_{i}^{t_{k}}-x_{0}^{t}}\right)
\end{array}\right.
\end{equation}
where $x_{0}^{t}$ and $y_{0}^{t}$ are the lateral and longitudinal coordinates of the ego vehicle (defined as the origin $\mathcal{O}$) at time $t$, respectively. $\rho_{i}^{t_{k}}$ is the Polar diameter relative to the origin $\mathcal{O}$, and $\theta_{i}^{t_{k}}$ is the orientation of the $i$th vehicle at time $t_{k}$.

\subsection{Multi-modal Probabilistic Maneuver Prediction}
\begin{figure}[htbp]
  \centering
\includegraphics[width=0.9\linewidth]{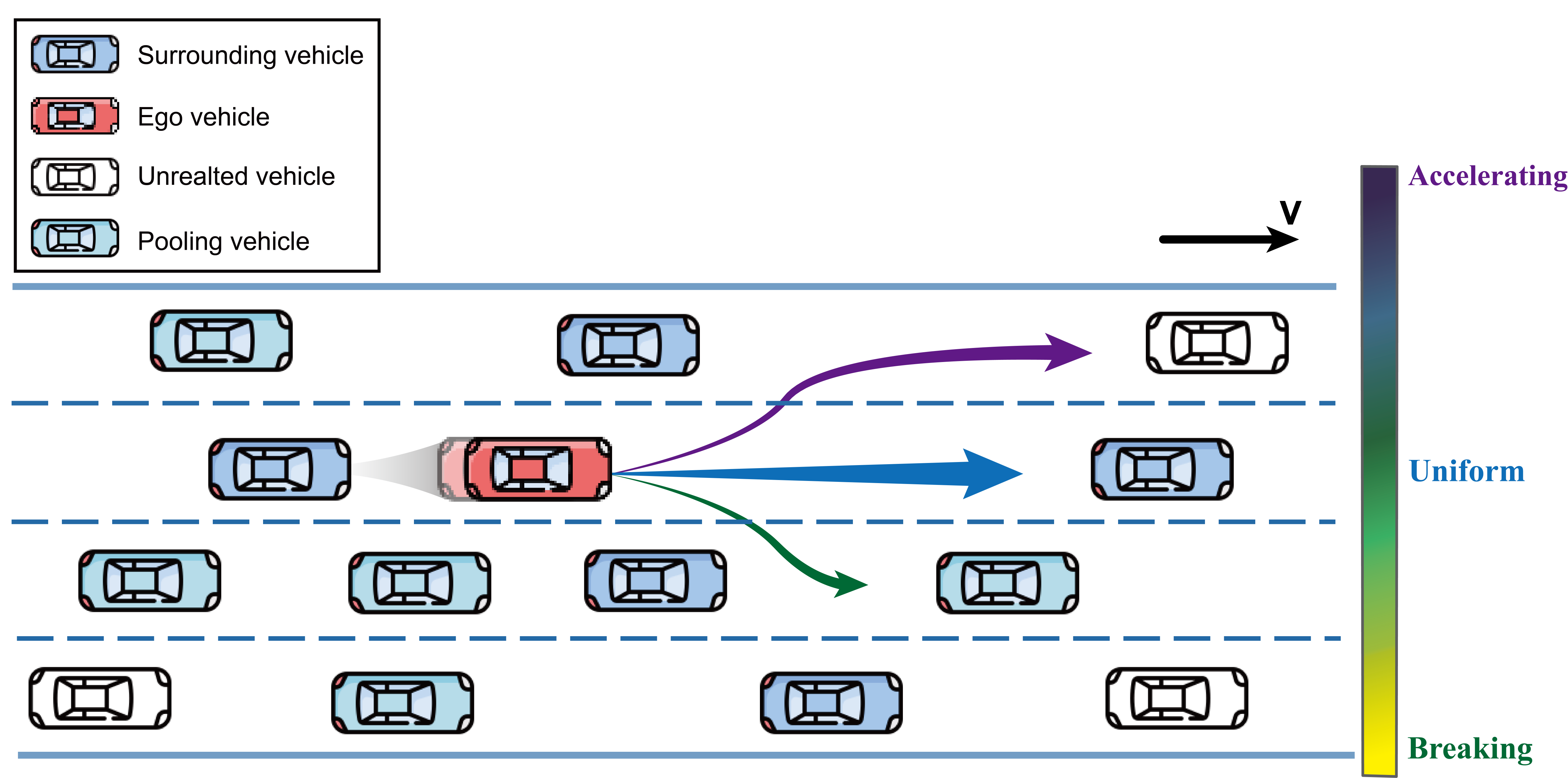} 
  \caption{Multi-modal maneuver prediction framework with corresponding probability outputs.}
  \label{manuver} 
\end{figure}

To account for the uncertainty and variability in the prediction, the multimodal prediction framework considers multiple potential maneuvers that the ego vehicle could perform and estimates the probability of each maneuver based on previous observations, as shown in Fig. \ref{manuver}. This not only provides multiple predictions but also quantifies the confidence level associated with each prediction. This is particularly beneficial for informed decision-making in response to anticipated maneuvers, as it allows AVs to account for the uncertainty inherent in the predictions.

To provide a formal description, the future trajectories can be hierarchically predicted in a Bayesian framework at two hierarchical levels: (1) At each time instant, the probability of different maneuvers $\bm{M}$ of the ego vehicle is determined; (2) Subsequently, the detailed trajectories of the vehicle conditional on each maneuver are generated within a predefined distributional form. In accordance with the characteristics of the driver's actions during driving, the possible maneuvers of the vehicles are decomposed into a combination of two distinctive sub-maneuvers, comprising the positioning-wise sub-maneuver $\bm{M}_{p}$ and the velocity-wise sub-maneuver $\bm{M}_{v}$. In addition, the position-wise sub-maneuver includes three discrete driver choices regarding position changes, namely left lane change, right lane change, and lane keeping. Meanwhile, the speed-wise sub-maneuver includes three distinct decisions, including accelerating, braking, and maintaining speed, which serve as branch categories.

Conditioned on the estimated maneuvers $\bm{M}$ in the first layer, the probability distribution of multimodal trajectory predictions is assumed to follow a Gaussian distribution: 
\begin{equation}
    P_{\bm{\Omega}} (\bm{Y}|\bm{M},\bm{X}) = N(\bm{Y}|\mu(X),\Sigma(X))
\end{equation}
where $\bm{\Omega}=\left[\Omega^{t+1}, \ldots,\Omega^{t+t_{f}}\right]$ are the estimable parameters of the distribution, and $\Omega^{t}=[\mu^{t},\Sigma^{t}]$ is the mean and variance of the distribution of predicted trajectory point at $t$. Correspondingly, in the second layer, the multimodal predictions are formulated as a Gaussian Mixture Model, i.e.,
\begin{equation}\label{eq.6}
    P(\bm{Y}|\bm{X})=\sum_{\forall i} P\left(M_{i}|\bm{X}\right) P_{\bm{\Omega}}\left(\bm{Y}|M_{i},\bm{X}\right)
\end{equation}
where $M_{i}$ is the $i$-th element in $\bm{M}$.

\section{Proposed Model}\label{Proposed Model}
\begin{figure*}[t]
  \centering
  \includegraphics[width=0.95\linewidth]{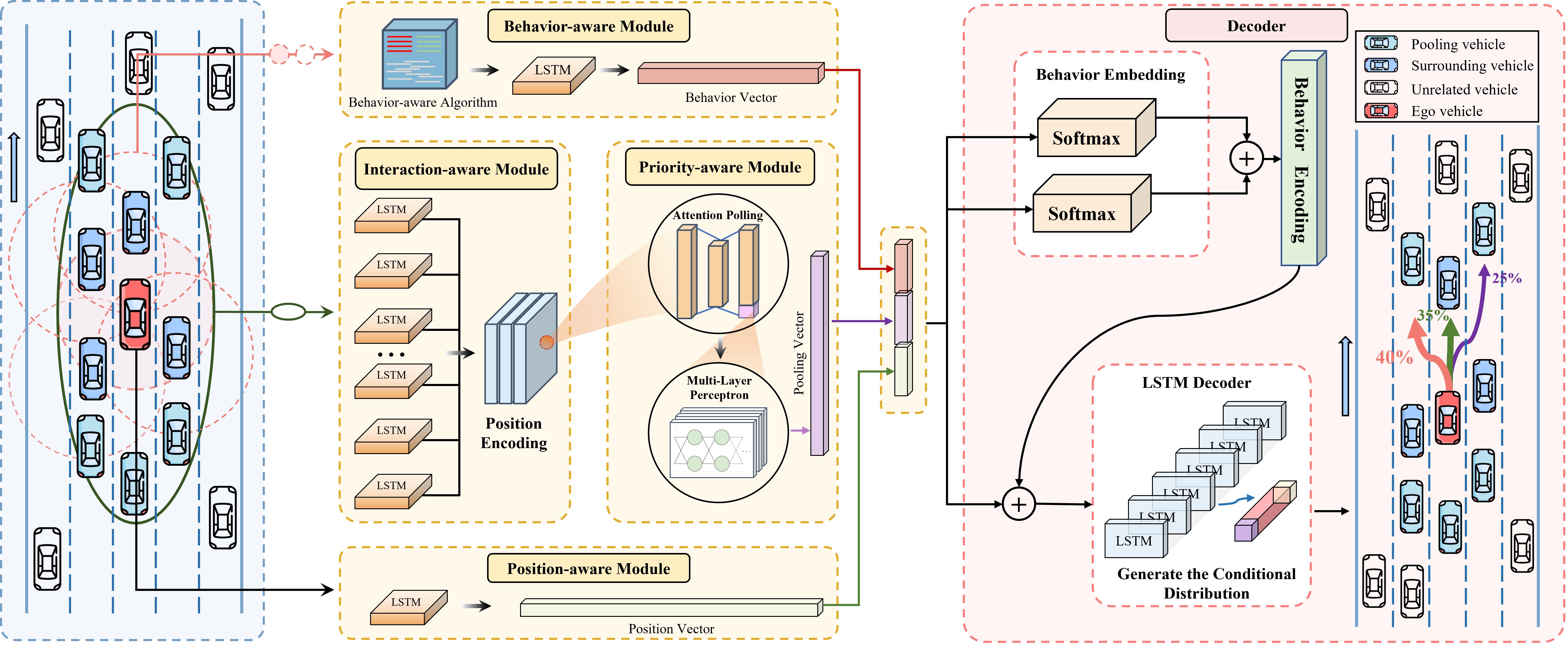} 
  \caption{Architecture of behavior-aware trajectory prediction model}
  \label{fig5} 
\end{figure*}
Fig. \ref{fig5} shows the architecture of BAT, which is upon the encoder-decoder framework with four modules to capture different aspects of behaviors and interactions between different agents, including behavior-aware, interaction-aware, priority-aware, and position-aware modules.

\subsection{Behavior-aware Module}\label{Behavior-aware Module_0}
The complex and dynamic nature of traffic scenarios presents significant challenges in interpreting and categorizing driver behavior. Unlike previous studies that categorize driver behavior into finite and human-defined classifications, we present a more flexible and adaptable solution, namely the behavior-aware module, by avoiding discrete behavior categories in favor of a continuous representation of behavioral information. Our behavior-aware module is motivated by the multi-policy decision-making (MPDM) framework for human drivers \citep{markkula2020defining} and integrates traffic psychology \citep{toghi2022social} using dynamic geometric graphs (DGGs) \citep{dall2002random} to model and evaluate human driving behavior. 

\subsubsection{Dynamic Geometric Graphs}
At time $t$, the graph $G^{t}$ can be given as follow:
\begin{equation}\label{eq.7-1}
{G}^{t} = \{V^{t},{E}^{t}\}
\end{equation}
where $V^{t}=\{{v}_{0}^{t},{v}_{1}^{t}\ldots,{v}_{n}^t\}$ is the set of nodes, ${v}_{i}^{t}$ is the $i$-th node representing the $i$-th vehicle, ${E^{t}} = \{{e_{0}^{t}},{e_{1}^{t}}\ldots,{e_{n}^{t}}\}$ is the set of undirected edges, and ${e_{i}^{t}}$ is the edge between the node ${v}_{i}^{t}$ and other vehicles that have potential influences on it. It is assumed that the interaction only exists when the nodes $v_{i}$ and $v_{j}$ are in close proximity to one another, or formally, the shortest distance between them, $d\left(v_{i}^{t}, v_{j}^{t}\right)$, is less than or equal to the predetermined distance threshold $r$. Therefore, we define 
\begin{equation}\label{eq.7-4}
 {e_{i}^{t}}= \{{v_{i}^{t}} {v_{j}^{t}} \mid(j \in  {N}_{i}^{t})\}
\end{equation}
where $ {N}_{i}^{t}=\left\{v_{j}^{t} \in V^{t}\backslash\left\{v_{i}^{t}\right\}\mid d\left( v_{i}^{t}, v_{j}^{t}\right) \leq r, i \neq j\right\}$.

Correspondingly, the symmetrical adjacency matrix $A^{t}$ of $G^{t}$ can be given as:
\begin{equation}\label{eq.8}
A^{t}(i, j)= \begin{cases}d\left(v_{i}^{t}, v_{j}^{t}\right) & \text { if } d\left(v_{i}^{t}, v_{j}^{t}\right)\leq{r}, i \neq j \\ 0 & \text {otherwise}\end{cases}
\end{equation}

\subsubsection{Centrality Measures}\label{Centrality Measures}
To more accurately capture the potential interactions between the observed traffic agents, we use centrality measures (degree, closeness, and eigenvector centrality measures) \cite{freeman1978centrality} as prior knowledge to portray and  describe driving behavior in DGGs, as shown in Table \ref{table_behavior}. 

\textbf{Degree Centrality.}
Degree centrality is characterized by the count of immediate connections a node has with other nodes within the graph. This concept intuitively suggests that a traffic agent with more connections is both more susceptible to the influences of other agents and more influential in shaping their actions. Formally,
\begin{equation}\label{eq.11}
     \mathcal{J}_{i}^{t}(D)= \left| \mathcal{N}_{i}^{t}\right|+\mathcal{J}_{i}^{t-1}(D)
\end{equation}
where $\left| \mathcal{N}_{i}^{t}\right|$ is the total number of elements in $\mathcal{N}_{i}^{t}$ at time $t$.

\textbf{Closeness Centrality.}
We propose that the closer a vehicle is to its surroundings, the higher its likelihood of interacting with adjacent vehicles. This idea is encapsulated by the closeness centrality metric, which gauges the ease of interaction and accessibility between a vehicle and its neighboring vehicles. Closeness centrality is determined using the shortest paths between the vehicle (node) and other vehicles in the traffic graph. This is achieved by summing the inverse of their distances. Formally,
\begin{equation}\label{eq.12}
\mathcal{J}_{i}^{t}(C)=\frac{\left| \mathcal{N}_{i}^{t}\right|-1}{\sum_{\forall v_{j}^{t} \in \mathcal{N}^{t}_{i}}d\left(v_{i}^{t}, v_{j}^{t}\right)}
\end{equation}
\begin{table*}[t]
  \centering
  \caption{Centrality measures and their interpretations}
  \resizebox{\linewidth}{!}{ 
    \begin{tabular}{ccc}
    \toprule
    Centrality Measures & \multicolumn{1}{c}{Magnitude (Original Measure)} & Gradient (1st Derivative) \\
    \midrule
    Degree & Agent's potential and capability for interaction in the traffic environment & Agent's sensitivity to traffic density variations \\
    \midrule
     Closeness & \multicolumn{1}{c}{Agent's significance in dynamic traffic scenarios} & Variation in agent's importance in dynamic traffic scenes \\
    \midrule
    Eigenvector & Extent of influence an agent exerts on others via direct and indirect interactions at time t & Agent's capacity to modify interactions in complex traffic scenarios \\
    \bottomrule
    \end{tabular}%
    }
  \label{table_behavior}%
\end{table*}%

\textbf{Eigenvector Centrality.}
In the context of understanding driver behavior, a vehicle's eigenvector centrality takes into account both its interactions with nearby vehicles and the influence of those interactions. Specifically, this metric integrates the vehicle's number of connections and the weight of the influence of connected vehicles. This helps to identify influential vehicles in a traffic context and their potential impact on other drivers. Formally,
\begin{equation}\label{eq.13}
\mathcal{J}_{i}^{t}(E)=\frac{ \sum_{\forall v_{j}^{t} \in \mathcal{N}^{t}_{i}}d\left(v_{i}^{t}, v_{j}^{t}\right)}{\lambda}
\end{equation}
where $\lambda$ is the eigenvalue. In addition, the Perron-Frobenius theorem states that for a non-negative matrix (such as the adjacency matrix in our case), there exists a positive eigenvector solution for the greatest eigenvalue of the matrix \cite{pillai2005perron}. This means that the eigenvector corresponding to the greatest eigenvalue of the adjacency matrix can be used to compute the eigenvector centrality measure of the nodes in the graph. 

\subsubsection{Behavior-aware Criterion}
The behavior-aware criterion is devised to mirror human-like trajectory predictions by leveraging the analytical properties of centrality measures. This aids in detecting and comprehending human driving behavior. By doing so, it removes the necessity for manual labeling, addressing issues like labeling non-continuous behaviors and choosing optimal time frames. Furthermore, this criterion effectively encapsulates continuous driving behaviors. Incorporating this with the Behavior Likelihood Estimate (BLE) and Behavior Intensity Estimate (BIE) refines prediction accuracy and dependability in fluctuating and intricate traffic conditions.

\textbf{Behavior Likelihood Estimate.} 
The BLE criterion quantifies behavior probabilities using time-based row derivatives, even without explicit behavior classifications. A higher probability of a behavior is indicated by prominent row derivatives and local extrema. For $v_{i}^{t}$ at time $t$, the BLE considering all three centrality measures is as follows:
{\begin{footnotesize}
\begin{equation}\label{eq.14}
\operatorname{\mathcal{I}}_{i}^{t}=\left[\left|\frac{\partial \mathcal{J}^{t}_{i}(D)}{\partial t}\right|,\left|\frac{\partial \mathcal{J}^{t}_{i}(C)}{\partial t }\right|,\left|\frac{\partial \mathcal{J}^{t}_{i}(E)}{\partial t }\right|\right]^{T}
\end{equation}
\end{footnotesize}
}
where $\left| \cdot  \right|$  denotes the absolute value operator. 

\textbf{Behavior Intensity Estimate.}
The BIE quantifies the potential impact intensity of a driving behavior on surrounding vehicles. It takes into account the duration of the behavior, with longer-lasting behaviors assumed to have a greater impact than those that are brief. 
The BIE for the node $v_{i}^{t}$ at time $t$ is build on top of BLE, and is defined as:
{\begin{footnotesize}
\begin{flalign}\label{eq.15}
\operatorname{{\mathcal{L}}}_{i}^{t}  =\left|\frac{\partial\mathcal{\operatorname{{\mathcal{I}}}}_{i}^{t}}{\partial t}\right|  
     = \left[\left|\frac{\partial^{2} \mathcal{J}_{i}^{t}(D)}{\partial^{2} t}\right|,\left|\frac{\partial^{2} \mathcal{J}_{i}^{t}(C)}{\partial^{2} t }\right|, \left|\frac{\partial^{2} \mathcal{J}_{i}^{t}(E)}{\partial^{2} t }\right|\right]^{T} 
\end{flalign}
\end{footnotesize}}

In summary, the BIE in conjunction with the BLE as prior knowledge provides a comprehensive understanding of individual driver behavior. This is achieved by segmenting each traffic scene into behavior-aware regions centered around observed agents of the ego vehicle. For these regions, behavioral features are extracted from the traffic agents, including contextual information (as shown in Fig. \ref{polar}). These features are then embedded and encoded frame by frame by an LSTM network to generate high-dimensional behavior vectors. By combining insights into the probability and intensity of the behavior, the overall impact on the surrounding traffic is determined. This infusion of human-like reasoning aligns with human perception and cognition, improving the accuracy and efficiency of trajectory prediction for AVs.  


\subsection{Interaction-aware Module}\label{Interaction-aware Pooling Module_0}
To capture and assemble interactions between the ego vehicle and its surrounding agents, we introduce an innovative interaction-aware pooling mechanism. Unlike conventional methods \citep{deo2018convolutional,chen2022intention,wang2023wsip}, which treat agents as fixed-size grid cells with Cartesian coordinates, our mechanism uses relative displacements in polar coordinates. This better adapts to complex non-regularized scenarios such as roundabouts and irregular intersections. Hence, we apply a hierarchical LSTM encoder and a position encoding layer to enhance the effective and efficient awareness and aggregation of interactions between the ego vehicle and its surrounding vehicles in the context of highly interactive driving scenarios. The LSTM encoder effectively captures interactions and dynamic motion between surrounding agents. At each discrete time step, the LSTM encoder learns the most recent $t_{h}$ frames of historical trajectory information for both the ego vehicle and its observed agents, embedding their positions in a temporal order. The hidden position states of these vehicles are then updated by the LSTM on a frame-by-frame basis, with the weights of the LSTM shared across all vehicles. The position data is then represented in polar coordinates and mapped through the position encoding layer to capture higher-order interactive information.

\subsection{Priority-aware Module}
The priority-aware module uses an attention mechanism layer to compute dynamic attention weight vectors for surrounding agents based on their higher-order interactive information. This attention mechanism \cite{vaswani2017attention} assigns weights that indicate their importance in predicting the ego vehicle's trajectory. These weight vectors express the relative importance of the agents and are used to weight higher-order interaction data in later stages. They are then fed into a multi-layer perceptron (MLP) to produce high-dimensional aggregate pooling vectors through a max-pooling layer.

\subsection{Position-aware Module}\label{Position Embedding Module} 
To further enhance the modeling of temporal dependencies and spatial relationships, this module employs a dedicated LSTM network to encode and learn the dynamic position of the ego vehicle. The historical trajectory of the ego vehicle is also represented in polar coordinates and subsequently embedded using an LSTM. This refinement augments the model's capability to capture the details of the agent's trajectory.

\subsection{Decoder}\label{Decoder Moudle} 
The position vector of the ego vehicle is integrated with additional information about the hidden pooling vectors and the high-dimensional behavior vector. This composite undergoes embedding by a softmax activation function (behavior embedding), followed by processing by an MLP (behavior encoding). Finally, the processed input is analyzed by the LSTM decoder, which generates a probability distribution over the possible future trajectories of the ego vehicle.

\section{Experiments}\label{Experiments}
We evaluate the effectiveness of our model using four datasets: NGSIM \cite{deo2018convolutional}, HighD \cite{highDdataset}, RounD \cite{rounDdataset}, and MoCAD.  These datasets, sourced from varied and intricate real-world traffic situations like highways, roundabouts, and urban locales, serve as a comprehensive testing ground. To gauge our model's precision, we employed the Root Mean Square Error (RMSE) metric. 

\subsection{Experimental Setup}
These data sets were partitioned into training, validation, and test sets using standard sampling. We refer to the complete test set as the \textit{overall} test set. The trajectories for the NGSIM, HighD, and MoCAD datasets were divided into 8-second intervals. The first 3 seconds served as the trajectory history ($t_h=3$) for input, and the following 5 seconds represented the ground truth ($t_f = 5$) for output. For the RounD dataset, the trajectories were divided into 6-second chunks with $t_h=2$ and $t_f = 4$.

To delve deeper into our model's performance, the NGISM dataset was further split based on distinct vehicular maneuvers, including no lane-change (\textit{keep}), on-ramp lane merging  (\textit{merge}), right lane-change (\textit{right}), and left lane-change (\textit{left}). This subset, termed the \textit{maneuver-based} test set, allowed for a more granular examination of our model's capabilities across different traffic actions.

\subsection{Training and Implementation Details}\label{Training}
Our model is trained to converge using an NVIDIA A100 40GB GPU.  We introduce the Negative Log-Likelihood criterion as a complement to the RMSE criterion in the loss function. Our model is based on a Gaussian Mixture Model (GMM) with a multimodal predictive structure. The encoder comprises behavior-aware, interaction-aware, and position-aware LSTMs with dimensions of 32, 64, and 64, respectively, while the LSTM decoder uses 128 state dimensions. Training is performed for 12 epochs with the Adam optimizer (lr=0.001, batch size=256). To enhance training stability, we employ the CosineAnnealingWarmRestarts scheduler and set a threshold distance of 25 feet, which is the average headway. Additionally, for maneuver-based models, evaluating the correctness of predicted trajectories concerning intended maneuvers is crucial. The NLL criterion proves valuable in assessing the model's ability to capture the underlying dynamics and constraints of the vehicle's motion during different maneuvers, ensuring reliable trajectory predictions.
To address the potentially detrimental effects of misclassifying maneuver types on trajectory prediction accuracy and robustness, we propose complementing the RMSE criterion with the NLL criterion in the loss function. This diversity loss term incentivizes the model to generate trajectory predictions consistent with intended maneuvers, promoting trajectory diversity.

\begin{table}[!htb]
  \centering
     \caption{Evaluation results for BAT and the baselines in the \textit{overall} test set over a different horizon. Note: RMSE (m) is the evaluation metric, where lower values indicate better performance, with some not specifying ('-'). Values in \textbf{bold} represent the best performance in each category.}\label{Table1}
     \setlength{\tabcolsep}{1mm}
   \resizebox{\linewidth}{!}{
    \begin{tabular}{c|cccccc}
    \toprule
    \multicolumn{1}{c}{\multirow{2}[2]{*}{Dataset}} & \multirow{2}[3]{*}{Model} & \multicolumn{5}{c}{Prediction Horizon (s)} \\
\cmidrule{3-7}    \multicolumn{1}{c}{} &       & 1     & 2     & 3     & 4     & 5 \\
    \midrule
    \multirow{12}[27]{*}{NGSIM} & S-LSTM \cite{alahi2016social} & 0.65  & 1.31  & 2.16  & 3.25  & 4.55  \\
          & S-GAN \cite{gupta2018social} & 0.57  & 1.32  & 2.22  & 3.26  & 4.40  \\
          & CS-LSTM \cite{deo2018convolutional} & 0.61  & 1.27  & 2.09  & 3.10  & 4.37  \\
          & MATF-GAN \cite{zhao2019multi}& 0.66  & 1.34  & 2.08  & 2.97  & 4.13  \\
          & NLS-LSTM \cite{messaoud2019non}& 0.56  & 1.22  & 2.02  & 3.03  & 4.30  \\
          &M-LSTM \cite{deo2018multi}& 0.58  & 1.26  & 2.12  & 3.24  & 4.66 \\
          &IMM-KF \cite{lefkopoulos2020interaction}& 0.58  & 1.36  & 2.28  & 3.37  & 4.55 \\
          &GAIL-GRU \cite{kuefler2017imitating}& 0.69  & 1.51  & 2.55  & 3.65  & 4.71 \\
          &MFP \cite{tang2019multiple}& 0.54  & 1.16  & 1.89  & 2.75  & 3.78  \\
        &DRBP\cite{gao2023dual}& 1.18  & 2.83  & 4.22  & 5.82  & - \\
          & DN-IRL \cite{fernando2019neighbourhood}& 0.54  & 1.02  & 1.91  & 2.43  & 3.76  \\
          & WSiP \cite{wang2023wsip}& 0.56  & 1.23  & 2.05  & 3.08  & 4.34  \\
          & CF-LSTM \cite{xie2021congestion}& 0.55  & 1.10  & 1.78  & 2.73  & 3.82  \\
          & MHA-LSTM \cite{messaoud2021attention}& 0.41  & 1.01  & 1.74  & 2.67  & 3.83  \\
          & HMNet \cite{xue2021hierarchical}& 0.50  & 1.13  & 1.89  & 2.85  & 4.04  \\
          & TS-GAN \cite{wang2022multi}& 0.60  & 1.24  & 1.95  & 2.78  & 3.72  \\
          & STDAN \cite{chen2022intention}& 0.39  & 0.96  & 1.61  & 2.56 & 3.67  \\
          & \textbf{BAT (25\%)} & 0.31  & 0.85  & 1.65  & 2.69  & 3.87  \\
          & \textbf{BAT} & \textbf{ 0.23 } & \textbf{ 0.81 } & \textbf{ 1.54 } & \textbf{ 2.52 } & \textbf{ 3.62 } \\
    \midrule
    \multirow{7}[35]{*}{HighD} &S-LSTM \cite{alahi2016social}& 0.22  & 0.62  & 1.27  & 2.15  & 3.41  \\
    &S-GAN \cite{gupta2018social}& 0.30  & 0.78  & 1.46  & 2.34  & 3.41  \\
    &WSiP \cite{wang2023wsip}& 0.20  & 0.60  & 1.21  & 2.07  & 3.14  \\
    &CS-LSTM(M) \cite{deo2018convolutional}& 0.23  & 0.65  & 1.29  & 2.18  & 3.37  \\
    &CS-LSTM \cite{deo2018convolutional}& 0.22  & 0.61  & 1.24  & 2.10  & 3.27  \\
    &MHA-LSTM \cite{messaoud2021attention}& 0.19  & 0.55  & 1.10  & 1.84  & 2.78  \\
    &MHA-LSTM(+f) \cite{messaoud2021attention}& 0.06  & 0.09  & 0.24  & 0.59  & 1.18  \\
    &NLS-LSTM \cite{messaoud2019non}& 0.20  & 0.57  & 1.14  & 1.90  & 2.91  \\
    &DRBP\cite{gao2023dual}& 0.41  & 0.79  & 1.11  & 1.40  & - \\
    &EA-Net \cite{cai2021environment} & 0.15  & 0.26  & 0.43  & 0.78  & 1.32  \\
    &CF-LSTM \cite{xie2021congestion}& 0.18  & 0.42  & 1.07  & 1.72  & 2.44  \\
    &STDAN \cite{chen2022intention}& 0.19  & 0.27  & 0.48  & 0.91  & 1.66  \\
    &iNATran (M) \cite{chen2022vehicle} & \textbf{0.04}  & \textbf{0.05}  & 0.21  & 0.54  & 1.11  \\
    &iNATran \cite{chen2022vehicle}& \textbf{0.04}  & \textbf{0.05}  & 0.21  & 0.54  & 1.10 \\
     & \textbf{BAT (25\%)} & 0.14  & 0.34  & 0.65  & 0.89  & 1.27  \\
    & \textbf{BAT} & 0.08  & 0.14  & \textbf{ 0.20 } & \textbf{ 0.44 } & \textbf{ 0.62 } \\
     \midrule
    \multirow{7}[22]{*}{RounD}     
    &S-LSTM \cite{alahi2016social} & 0.94  & 1.82  & 3.43  & 5.21  & - \\
    &S-GAN \cite{gupta2018social} & 0.72  & 1.57  & 3.01  & 4.78  & -  \\
    &CS-LSTM(M) \cite{deo2018convolutional}& 0.74  & 1.43  & 2.44  & 4.21  & -  \\
    &CS-LSTM \cite{deo2018convolutional} & 0.71  & 1.21  & 2.09  & 3.92  & -  \\
    &MATH \cite{hasan2021maneuver} & 0.38  & 0.80  & 1.76  & 3.08  & - \\
    &MHA-LSTM \cite{messaoud2021attention} & 0.62  & 0.98  & 1.88  & 3.65  & -  \\
    &MHA-LSTM(+f) \cite{messaoud2021attention} & 0.51  & 0.91  & 1.80  & 3.57  & -  \\
    &NLS-LSTM \cite{messaoud2019non} & 0.62  & 0.96  & 1.91  & 3.48  & - \\
    &WSiP \cite{wang2023wsip} & 0.52  & 0.99  & 1.88  & 3.07  & -  \\
    &CF-LSTM \cite{xie2021congestion} & 0.51  & 0.87  & 1.79  & 3.14  & -  \\
    &STDAN \cite{chen2022intention} & 0.35  & 0.77  & 1.74  & 2.92  & -  \\
     & \textbf{BAT (25\%)} & 0.32  & 0.72  & 1.99  & 3.12  & -  \\
    & \textbf{BAT} & \textbf{0.23}  & \textbf{0.55}  & \textbf{1.43}  & \textbf{2.46}  & -  \\
     \midrule
    \multirow{7}[18]{*}{MoCAD}
    &S-LSTM \cite{alahi2016social} & 1.73  & 2.46  & 3.39  & 4.01  & 4.93 \\
    &S-GAN \cite{gupta2018social} & 1.69  & 2.25  & 3.30  & 3.89  & 4.69  \\
    &CS-LSTM(M) \cite{deo2018convolutional}& 1.49  & 2.07  & 3.02  & 3.62  & 4.53   \\
    &CS-LSTM \cite{deo2018convolutional} & 1.45  & 1.98  & 2.94  & 3.56  & 4.49  \\
    &MHA-LSTM \cite{messaoud2021attention} & 1.25  & 1.48  & 2.57  & 3.22  & 4.20  \\
    &MHA-LSTM(+f) \cite{messaoud2021attention} & 1.05  & 1.39  & 2.48  & 3.11  & 4.12  \\
    &NLS-LSTM \cite{messaoud2019non} & 0.96  & 1.27  & 2.08  & 2.86  & 3.93\\
    &WSiP \cite{wang2023wsip} & 0.70  & 0.87  & 1.70  & 2.56  & 3.47  \\
    &CF-LSTM \cite{xie2021congestion} & 0.72  & 0.91  & 1.73  & 2.59  & 3.44 \\
    &STDAN \cite{chen2022intention} & 0.62  & 0.85  & 1.62  & 2.51  & 3.32  \\
     & \textbf{BAT (25\%)}& 0.65  & 0.99  & 1.89  & 2.81  & 3.58 \\
    & \textbf{BAT} & \textbf{0.35}  & \textbf{0.74}  & \textbf{1.39}  & \textbf{2.19}  & \textbf{2.88}  \\
    \bottomrule
    \end{tabular}%
  \label{tab:addlabel}%
  }
\end{table}%

\subsection{Experimental Results} We evaluated our model against various SOTA trajectory prediction methods from 2016 to 2023. The results, displayed in Table \ref{Table1}, highlight our model's significant advancements in trajectory prediction over current SOTA baselines. Using RMSE as the evaluation metric, our model surpasses recent baselines (2021-2023) by 2.6\% for short-term predictions (1s-3s) and reduces prediction error by 56.7\% for long-term predictions (4s-5s) on the NGSIM dataset. On the HighD dataset, known for its superior data volume and precision, our model significantly outperforms most baselines, showing improvements of 62.7\% and 43.6\% compared to STDAN and iNATran, respectively, over a 5-second horizon.

The strengths of BAT become more evident in complex scenarios, like urban streets and unstructured roads (RounD and MoCAD datasets). Here, our model consistently surpasses current SOTA baselines, with accuracy gains between 17.8\%-75.5\% on RounD and 12.7\%-79.8\% on MoCAD. Such improvements underscore the significance of factoring in driving behavior and our relative distance pooling mechanism, especially in dense traffic scenarios. For scalability testing, even when our model was trained on just 25\% of the training data, it still managed to outperform most baselines, indicating a potential reduction in data needs for training AVs in challenging contexts. 

We also conducted tests on the \textit{maneuver-based} test set, as detailed in Table \ref{Table_manuver}.
Specifically, in the \textit{merge} and \textit{right} test subsets, our model achieves significantly lower RMSE values than the SOTA baselines, demonstrating an improvement of at least 10.1\% for a prediction horizon of 5 seconds, which could significantly mitigate the risk of traffic accidents. Moreover, our model shows remarkable improvement in the \textit{keep} and \textit{left} test subsets, highlighting its robustness and effectiveness in accurately predicting future vehicle trajectories in various driving scenarios and maneuvers.

Overall, our findings affirm our model's capability and efficiency in predicting vehicle trajectories for AVs.

\subsection{Ablation Studies}
Table \ref{Table4} presents an analysis of four critical components: polar coordinates, behavior-aware, interaction-aware, and priority-aware modules. We tested five models: Model A (using Cartesian coordinates), Model B (excluding the behavior-aware module), Model C (excluding the interaction-aware module), Model D (excluding the priority-aware module), and Model E (with all components).

On evaluating against the NGSIM and RounD datasets, all stripped-down versions (A-D) underperformed compared to the comprehensive Model E. Notably, the integration of interaction-aware and priority-aware modules significantly boosted performance, underlining their importance in enhancing prediction accuracy.

\begin{table}[htb]
  \centering
  \caption{Evaluation results for the proposed model and the baselines in maneuver-base test set for NGSIM dataset.}
   \resizebox{\linewidth}{!}{
    \begin{tabular}{ccccccccccc}
    \toprule
    Dataset & \multicolumn{5}{c}{\textit{keep}}     & \multicolumn{5}{c}{\textit{merge}} \\
    \cmidrule(r){1-6} \cmidrule(r){7-11} 
    \multirow{2}[1]{*}{Model} & \multicolumn{5}{c}{Horizon (s)}       & \multicolumn{5}{c}{Horizon (s)} \\
\cmidrule(r){2-6} \cmidrule(r){7-11}          & 1     & 2     & 3     & 4     & 5     & 1     & 2     & 3     & 4     & 5 \\
    S-LSTM \cite{alahi2016social} & 0.35  & 1.01  & 1.81  & 2.82  & 4.15  & 0.81  & 1.31  & 2.51  & 4.01  & 5.78 \\
    S-GAN \cite{gupta2018social} & 0.36  & 1.01  & 1.81  & 2.83  & 4.15  & 0.71  & 1.32  & 2.53  & 4.11  & 5.97 \\
    CS-LSTM \cite{deo2018convolutional} & 0.34  & 0.98  & 1.75  & 2.77  & 4.06  & 0.61  & 1.34  & 2.58  & 4.12  & 5.94 \\
    MATF-GAN \cite{zhao2019multi}& 0.37  & 1.11  & 1.74  & 2.66  & 3.91  & 0.53  & 1.41  & 2.56  & 3.97  & 5.52 \\
    WSiP \cite{wang2023wsip} & 0.32  & 0.89  & 1.58  & 2.51  & 3.59  & 0.40  & 1.18  & 2.41  & 3.72  & 5.16 \\
    HMNet \cite{xue2021hierarchical} & 0.31  & 0.83  & 1.56  & 2.51  & 3.68  & 0.34  & 1.17  & 2.32  & 3.63  & 5.20 \\
    STDAN \cite{chen2022intention}& 0.28  & 0.85  & 1.52  & 2.53  & 3.49  & 0.28  & 1.19  & 2.21  & 3.67  & 4.95 \\
    \textbf{BAT (25\%)} & 0.28  & 0.86  & 1.54  & 2.52  & 3.73  & 0.31  & 0.95  & 1.95  & 3.31  & 4.98 \\
    \textbf{BATaj} & \textbf{0.23}  & \textbf{0.81}  & \textbf{1.49}  & \textbf{2.44}  & \textbf{3.66}  & \textbf{0.25}  & \textbf{0.89}  & \textbf{1.83}  & \textbf{3.04}  & \textbf{4.45} \\
    \midrule
          &       &       &       &       &       &       &       &       &       &  \\ 
    \midrule
    Dataset & \multicolumn{5}{c}{\textit{left}}     & \multicolumn{5}{c}{\textit{right}} \\
    \cmidrule(r){1-6} \cmidrule(r){7-11} 
    \multirow{2}[3]{*}{Model} & \multicolumn{5}{c}{Horizon (s)}       & \multicolumn{5}{c}{Horizon (s)} \\
    \cmidrule(r){2-6} \cmidrule(r){7-11}         & 1     & 2     & 3     & 4     & 5     & 1     & 2     & 3     & 4     & 5 \\
    S-LSTM \cite{alahi2016social}& 0.77  & 1.68  & 3.04  & 4.67  & 6.59  & 0.69  & 1.97  & 3.81  & 6.17  & 9.09 \\
    S-GAN \cite{gupta2018social}& 0.66  & 1.68  & 3.11  & 4.85  & 6.87  & 0.72  & 1.97  & 3.91  & 6.32  & 9.23 \\
    CS-LSTM \cite{deo2018convolutional}& 0.54  & 1.63  & 3.01  & 4.71  & 6.63  & 0.61  & 2.01  & 3.97  & 6.48  & 9.48 \\
    MATF-GAN \cite{zhao2019multi}& 0.61  & 1.72  & 3.02  & 4.62  & 6.34  & 0.56  & 1.88  & 3.90   & 6.07  & 9.01 \\
     WSiP \cite{wang2023wsip}& 0.41  & 1.46  & 2.82  & 4.42  & 6.22  & 0.52  & 1.61  & 3.60  & 5.78  & 8.45 \\
    HMNet \cite{xue2021hierarchical}& 0.41  & 1.31  & 2.87  & 4.47  & 6.33  & 0.49  & 1.62  & 3.47  & 5.87  & 8.59 \\
    STDAN \cite{chen2022intention}& 0.35  & 1.33  & 2.84  & 4.51  & 5.97  & 0.38  & 1.49  & 3.46  & 5.87  & 7.93 \\
    \textbf{BAT (25\%)} & 0.43  & 1.24  & 2.43  & 4.01  & 5.91  & 0.47  & 1.41  & 3.09  & 5.19  & 7.87 \\
    \textbf{BAT} & \textbf{0.33}  &\textbf{1.07}  & \textbf{2.24}  & \textbf{3.73}  & \textbf{5.51}  & \textbf{0.31}  & \textbf{1.36}  & \textbf{2.96}  & \textbf{5.15}  & \textbf{6.78} \\
    \bottomrule
    \end{tabular}%
    }
    \label{Table_manuver}
\end{table}%

\begin{table}[htbp]
  \centering
  \caption{Ablation results for different models on NGSIM and RounD datasets (Evaluation metric: RMSE (m))}
\setlength{\tabcolsep}{6mm}
\resizebox{\linewidth}{!}{
    \begin{tabular}{c|cccccc}
    \toprule
    \multicolumn{1}{c}{\multirow{2}[4]{*}{Dataset}} & \multirow{2}[4]{*}{Time (s)} & \multicolumn{5}{c}{ Model ($\Delta${Method} E)} \\
\cmidrule{3-7}    \multicolumn{1}{c}{} &       & A     & B     & C     & D     & E \\
    \midrule
    \multirow{5}[2]{*}{NGSIM} & 1     
    &0.27  & 0.30  & 0.27  & 0.28  & \textbf{0.23} \\
          & 2     & 0.86  & 0.89  & 0.85  & 0.87  & \textbf{0.81} \\
          & 3     & 1.63  & 1.68  & 1.60  & 1.63  & \textbf{1.54} \\
          & 4     & 2.65  & 2.68 & 2.62  & 2.64  & \textbf{2.52} \\
          & 5     & 4.02  & 4.08  & 3.97  & 3.97  & \textbf{3.62} \\
    \midrule
    \multirow{5}[1]{*}{RounD} & 1     & 0.76  & 0.55 & 0.44  & 0.35 & \textbf{0.23} \\
          & 2     & 0.94  & 0.83  & 0.76   & 0.72  & \textbf{0.55} \\
          & 3     & 1.87  & 1.72  & 1.63  & 1.54  & \textbf{1.43} \\
          & 4     & 3.02  & 2.82  & 2.76  & 2.68  & \textbf{2.46} \\
    \bottomrule
    \end{tabular}%
    }
 \label{Table4}
\end{table}%

The behavior-aware module's inclusion significantly enhanced performance by capturing dynamic vehicular interactions, vital for accurate trajectory prediction. By factoring in surrounding vehicles' behavior, BAT predicts the ego vehicle's trajectory more insightfully. This mirrors human decision-making, where actions and intentions of other agents, including vehicles, shape trajectory predictions \cite{baron2000thinking}.

Furthermore, adopting the polar coordinate system in Model C outperformed the Cartesian approach, especially in roundabout environments like RounD. This aligns with studies on human perception, suggesting people process goal-relevant information distinctively \cite{todd2000precis}. The polar system better reflects human cognition of spatial vehicular relationships, emphasizing the significance of both behavioral and spatial considerations in trajectory prediction.

\subsection{Ablation Study for Distance Threshold}
We further explored the effect of the proper setting of the threshold distance $d\left( v_{i}^{t}, v_{j}^{t}\right)$ of two vehicles on model performance. To this end, we conducted experiments using three different distance threshold parameters: 0 feet (0 meters), 25 feet (7.62 meters), and 50 feet (15.24 meters). The results, depicted in Fig. 2, reveal that when the predefined distance threshold parameter r is set to 0 feet (0 meters), the interaction between vehicles is disregarded, leading to a significant deterioration in the model's ability to predict trajectories. Conversely, when $r=25$ feet, the model's performance for trajectory prediction is significantly improved, while it declines at $r=50$ feet. 
 
These findings align with research on human attention and decision-making, which suggests that humans tend to prioritize information that is relevant to their current goals and that is within their immediate visual field \citep{kahneman1973attention}. This is likely because the brain has limited processing resources and must prioritize information in order to make efficient decisions. In the context of driving, this would mean that human drivers are more likely to pay attention to and be influenced by the movements of nearby vehicles. By setting the distance threshold to 25 feet (7.62 meters), the model is able to capture the complex and dynamic interactions between vehicles that are most likely to impact the ego vehicle's trajectory, in a manner that is consistent with human attention and decision-making.
\begin{figure}[htbp]
  \centering
\includegraphics[width=0.9\linewidth]{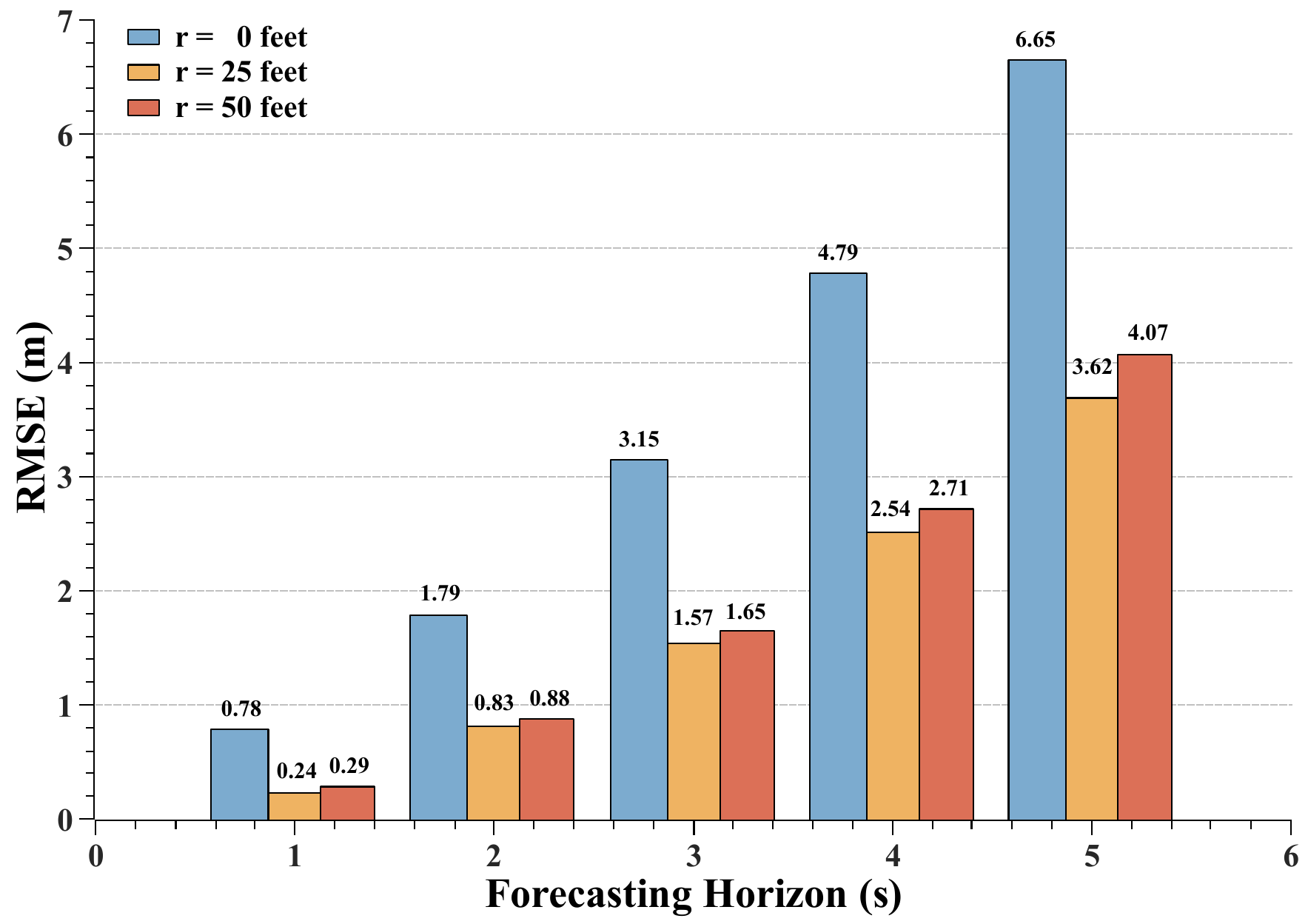} 
  \caption{Ablation results for different distance thresholds}
  \label{trajectory} 
\end{figure}

\begin{figure}[t]
  \centering
\includegraphics[width=\linewidth]{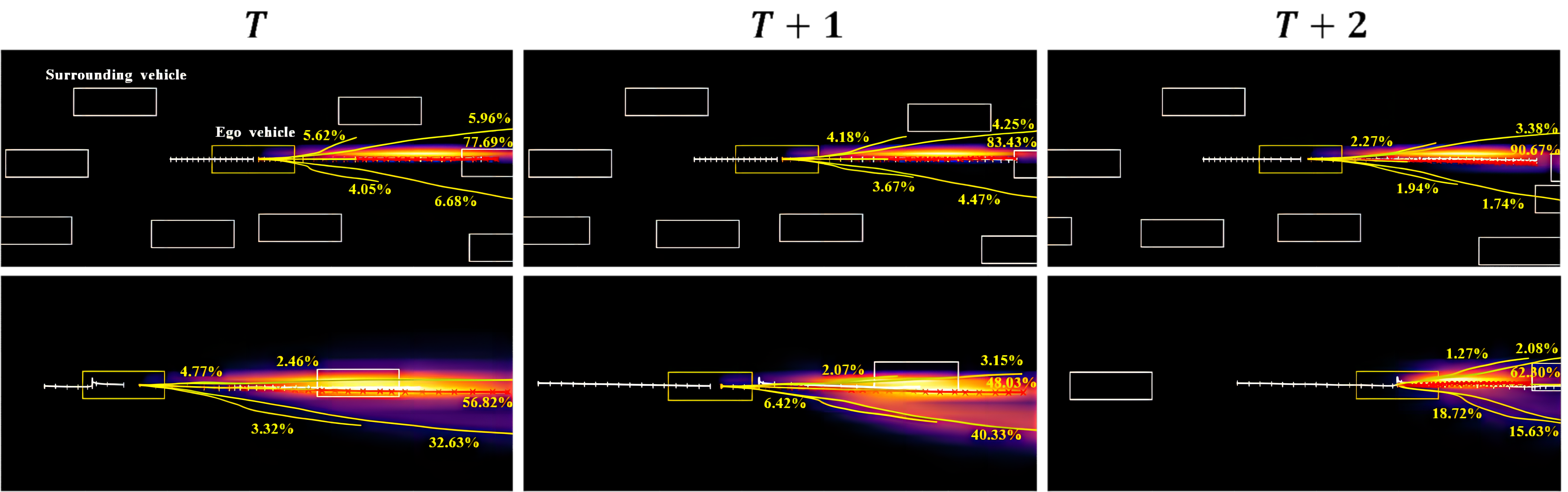} 
  \caption{Multi-modal probabilistic prediction of the ego vehicle. The heat maps depict the Gaussian Mixture Model of predicted outcomes at each time step, where brighter colors indicate higher probabilities.}
  \label{fig1} 
\end{figure}

\begin{figure}[t]
  \centering
\includegraphics[width=0.95\linewidth]{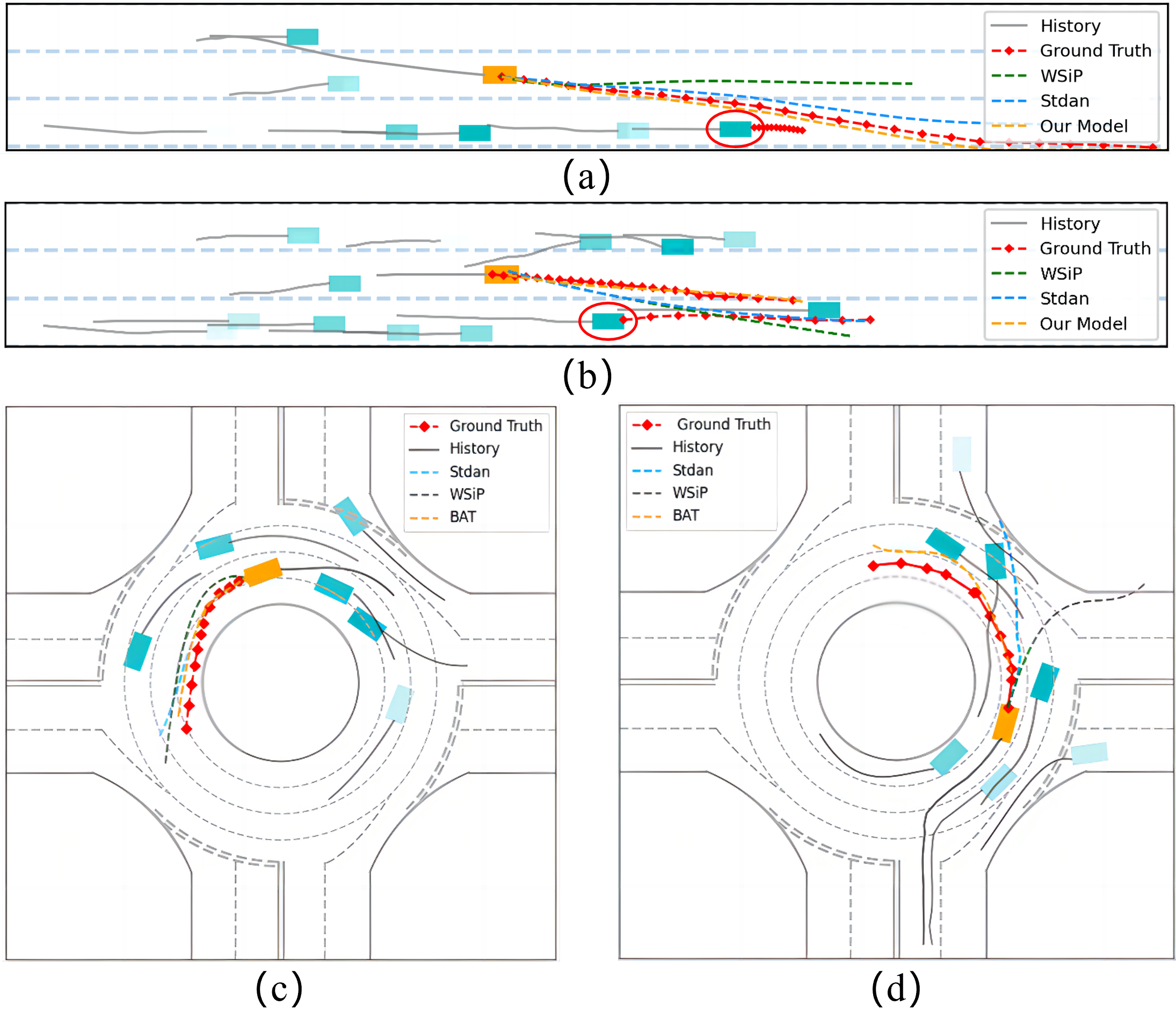} 
  \caption{Visualizations and heat maps selected from the NGSIM (a-b) and RounD (c-d) datasets. The target vehicle is depicted in orange, while its surrounding vehicles are shown in blue. The darkness of the blue color indicates the higher importance weight of the surrounding vehicle.}
  \label{fig_case} 
\end{figure}

\subsection{Intuition and Interpretability Analysis}
Figure \ref{fig1} illustrates our model's multi-modal probabilistic prediction performance on the NGSIM dataset. To further underscore the prowess of BAT, we visually dissect its prediction outcomes across diverse scenarios in Fig. \ref{fig_case}. For the sake of clarity, we spotlight solely the trajectories deemed most probable for the ego vehicle in each context. We meticulously chose two demanding driving situations: transitioning into the right lane (Fig.\ref{fig_case} (a-b)) and maneuvering through a roundabout (Fig.\ref{fig_case} (c-d)). Intriguingly, the heat maps vividly unveil a direct relationship between the proximity of the ego to neighboring vehicles and their respective significance. This exposes pronounced social interplay among the nearby agents. In Fig. \ref{fig_case} (a), the bottom-most vehicle with the red circle exhibits friendly driving behavior, intuitively creating ample space for the ego vehicle's lane transition. On the flip side, the vehicle with the red circle  in Fig. \ref{fig_case} (b) manifests aggressive driving tendencies, potentially accelerating to impede the ego vehicle's lane merge. Herein lies the genius of BAT’s behavior-aware module: it discerns driver personas, predicting the ego vehicle's inability to seamlessly merge, aligning impeccably with the ground truth. Conversely, models bereft of this driving behavior consideration falter, such as Stdan and WSiP, deviating significantly from the actual trajectory.

In addition, BAT captures the influence of agents even from non-adjacent lanes, attributing this to their distinct driving behavior—a facet frequently sidestepped in conventional studies. To sum up, BAT doesn't just predict; it observes, interprets, and decides like a human. By mirroring human decision-making, BAT offers a promising leap toward autonomous driving that's both accurate and reliable.

\section{Conclusion}\label{Conclusion} Predicting the trajectories of surrounding vehicles with a high degree of accuracy is a fundamental challenge that must be addressed in the quest for full AVs. To address this challenge, we propose a behavior-aware modular model with four components: behavior-aware, interaction-aware, priority-aware, and position-aware modules. Our model outperforms current SOTA baselines in terms of prediction accuracy and efficiency on the NGSIM, HighD, RounD, and MoCAD datasets, even when trained on 25\% training set, demonstrating its robustness, applicability, and potential to reduce training data requirements for AVs in challenging or unusual situations such as corner cases, roundabouts. 

\section*{Acknowledgement}
This research is supported by the Science and Technology Development Fund of the Macau SAR (File No. 0021/2022/ITP, 0081/2022/A2, 0015/2019/AKP, SKL-IoTSC(UM)-2021-2023/ORP/GA08/2022, SKL-IoTSC(UM)-2024-2026/ORP/GA06/2023) and the University of Macau (SRG2023-00037-IOTSC). Part of this research is carried out at SICC with support from SKL-IOTSC, University of Macau. For any correspondence regarding this work, please contact Dr. Zhenning Li (zhenningli@um.edu.mo) and Dr. Chengzhong Xu (czxu@um.edu.mo).

\bibliography{aaai24}

\end{document}